\documentclass[11pt]{article}

\usepackage[T1]{fontenc}
\usepackage{array}
\usepackage[table]{xcolor}
\usepackage{siunitx}
\usepackage{multirow}
\usepackage{etoolbox}
\usepackage{newtxtext}
\usepackage{newtxmath}
\usepackage{microtype}
\usepackage[margin=1in]{geometry}
\usepackage{amsmath,amsfonts}
\usepackage{booktabs}
\usepackage{longtable}
\usepackage{placeins}
\usepackage{natbib}
\usepackage{xurl}
\usepackage{titlesec}
\usepackage[hidelinks]{hyperref}
\newcommand{\ParGLUSD}[1]{{\scriptsize\(\pm\,#1\)}}
\newcommand{\ParGLUTableStyle}{\small\renewcommand{\arraystretch}{1.05}\setlength{\tabcolsep}{3pt}}
\definecolor{GrpBg}{HTML}{F2F4F7}
\robustify\bfseries
\hypersetup{pdftitle={Partial Gating Improves Language Modeling in MoE Transformers},pdfauthor={Yuting Ge}}

\titlespacing*{\section}{0pt}{1.15ex plus .2ex}{.55ex}
\titlespacing*{\subsection}{0pt}{.85ex plus .15ex}{.35ex}
\titlespacing*{\subsubsection}{0pt}{.65ex plus .1ex}{.25ex}
\title{Partial Gating Improves Language Modeling in MoE Transformers}
\author{Yuting Ge\\[0.4ex]\normalsize\href{mailto:yutingge3-c@my.cityu.edu.hk}{yutingge3-c@my.cityu.edu.hk}}
\date{}

\begin{document}

\maketitle

\begin{abstract}
Multiplicative gating improves Transformer feed-forward networks (FFNs), but its additional projection reduces intermediate representation width under a fixed parameter budget. This trade-off is particularly important in Mixture-of-Experts (MoE) Transformers, where model capacity is distributed across individual experts. We introduce Partial Gated Linear Units (ParGLU), a simple expert architecture that combines parallel SwiGLU and squared-ReLU branches without increasing parameters or modifying expert routing.

We conduct 100 pretraining runs on two corpora, spanning 30M to 1.03B parameters and approximately 247 billion training tokens. ParGLU outperforms SwiGLU across all three seeds at 75M and 152M and all 11 configurations in the scaling experiments. Controlled ablations reveal the importance of gating ratio, activation choice, expert width, and expert type. ParGLU achieves training throughput comparable to SwiGLU, even surpassing it in several configurations. Full fine-tuning further yields lower perplexity on all three downstream test sets.

Token-level analyses reveal pronounced improvements on low-frequency tokens and high-entropy predictions. At 1.03B, ParGLU naturally produces 85.75\% exact zero activations in its squared-ReLU branch, enabling a theoretical reduction of 21.44\% in expert projection MACs. At nearly identical theoretical computation costs, ParGLU achieves 2.30\% lower perplexity than magnitude-pruned SwiGLU. These results establish partial gating as a simple and effective design choice for MoE Transformers.
\end{abstract}

\section{Introduction}

Feed-forward networks (FFNs) are a fundamental component of Transformers. The original Transformer adopted a two-projection FFN with ReLU activation and an intermediate width of $4d$, where $d$ denotes the model dimension \citep{Vaswani2017}. GELU later became a widely used alternative while retaining the same structure \citep{HendrycksGimpel2016}. Gated variants such as SwiGLU introduced a third projection to enable multiplicative interactions between independently learned representations \citep{Shazeer2020}. Under an equivalent parameter and principal matrix computation budget, this additional projection reduces the intermediate width from $4d$ to approximately $8d/3$. In contrast, squared ReLU (ReLU$^2$) retains the two-projection structure while introducing a squared activation \citep{So2021}. These architectures illustrate a fundamental trade-off between multiplicative gating and intermediate representation width.

Mixture-of-Experts (MoE) Transformers bring this trade-off into sharper focus. By distributing FFN parameters across sparsely activated experts, MoE architectures enable greater total model capacity without a proportional increase in computation per token. Recent studies have explored expert granularity, expert dimensions, and active parameter allocation as central design choices in MoE architectures \citep{Dai2024DeepSeekMoE,Ludziejewski2024Scaling,pmlr-v267-ludziejewski25a,Li2026SlicingDicing}. Yet while existing work focuses on routing tokens and partitioning capacity across experts, conventional SwiGLU experts spend their entire parameter budget on gated transformations, paying the full width penalty of a third projection.

We break this all-or-nothing convention with Partial Gated Linear Units (ParGLU), a simple expert architecture that combines SwiGLU and ReLU$^2$ in two independently parameterized parallel branches. Their outputs are simply added to produce the expert output. By reallocating a portion of the expert parameter budget to non-gated transformations, ParGLU retains multiplicative gating while expanding the total intermediate representation width. This modification requires no additional parameters, mixing gates, or changes to expert routing.

We evaluate ParGLU through 100 pretraining runs on SmolLM and FineWebEdu, spanning model sizes from 30M to 1.03B parameters and approximately 247 billion training tokens. Despite its simplicity, ParGLU consistently improves language modeling at 75M and 152M across three random seeds and outperforms SwiGLU in all 11 FineWebEdu configurations across three expert granularities. Controlled ablations demonstrate the benefits of partial gating across multiple allocation ratios and reveal the influence of expert width and expert type. Full fine-tuning of the 1.03B models further demonstrates improvements in perplexity across all three downstream test sets and in seven of nine ROUGE comparisons. ParGLU retains at least 96.7\% of SwiGLU training throughput in nine of the eleven scaling configurations, surpassing SwiGLU in three of them.

Beyond these performance improvements, ParGLU exhibits pronounced advantages on low-frequency tokens and predictions with high entropy. Its ReLU$^2$ branch also introduces native activation sparsity without additional pruning or sparsity regularization. In the 1.03B model, 85.75\% of ReLU$^2$ activations are exactly zero, enabling a theoretical reduction of 21.44\% in expert projection MACs. At nearly identical theoretical computation costs, native ParGLU achieves 2.30\% lower perplexity than magnitude-pruned SwiGLU.

Our contributions are threefold:
\begin{itemize}
    \item We propose ParGLU, a simple and effective expert architecture that introduces partial gating under a fixed parameter budget.
    \item We establish the effectiveness of partial gating across model scales and expert granularities, identifying key factors in its application to MoE experts.
    \item We demonstrate that partial gating improves the quality--computation trade-off through native activation sparsity.
\end{itemize}

\section{Related Work}

\subsection{Gated and Non-Gated Feed-Forward Networks}

\citet{Dauphin2017} introduced Gated Linear Units (GLUs) for language modeling. \citet{Shazeer2020} later investigated GLU variants in Transformer FFNs, including SwiGLU. Squared ReLU (ReLU$^2$) offers a non-gated alternative using a single intermediate projection \citep{So2021}. \citet{Arefin2026} analyzed the expressivity of tied and untied multiplicative interactions. Their work highlights the distinct representational properties of ReLU$^2$ and SwiGLU. Under a fixed parameter budget, SwiGLU uses independent projections for multiplicative gating, while ReLU$^2$ retains a wider intermediate representation. ParGLU combines these transformations.

\subsection{Expert Granularity and Capacity Allocation in MoE}

Expert granularity is an important design choice in MoE Transformers. DeepSeekMoE introduced fine-grained expert segmentation and shared experts \citep{Dai2024DeepSeekMoE}. \citet{Ludziejewski2024Scaling} studied how expert granularity affects scaling under fixed computational budgets. Their subsequent work extended MoE scaling laws to include active parameters and memory constraints \citep{pmlr-v267-ludziejewski25a}. More recently, \citet{Li2026SlicingDicing} systematically investigated expert dimensions and their relationship with active parameter budgets. These studies establish expert size and capacity allocation as central design considerations in MoE architectures. ParGLU focuses on a complementary problem: allocating the parameter budget within each expert between gated and non-gated transformations.

\section{Method}

\subsection{Partial Gated Linear Units (ParGLU)}

Partial Gated Linear Units (ParGLU) allocate the parameter budget of a Mixture-of-Experts (MoE) expert between two parallel branches: a gated SwiGLU branch and a non-gated squared-ReLU branch. Given an input $X\in\mathbb{R}^{C}$, the two branches are defined as

\begin{equation}
\mathcal{S}(X)=\left[\operatorname{SiLU}(XW_G)\odot XW_V\right]W_{O,S},
\end{equation}

\begin{equation}
\mathcal{R}(X)=\left[\operatorname{ReLU}(XW_R)\right]^2W_{O,R}.
\end{equation}

Their outputs are added to produce the expert output:

\begin{equation}
\operatorname{ParGLU}(X)=\mathcal{S}(X)+\mathcal{R}(X).
\end{equation}

The two branches are independently parameterized. ParGLU can replace the FFN in shared or routed experts without modifying the underlying MoE routing mechanism.

\subsection{Parameter Budget Allocation}

Let $C$ denote the backbone dimension and $H$ the intermediate width of a conventional SwiGLU expert. A bias-free SwiGLU expert requires $3CH$ parameters and principal dense projection MACs per token. For ParGLU, let $S$ and $R$ denote the intermediate widths of the gated and non-gated branches, respectively. Preserving the original budget requires

\begin{equation}
3CS+2CR=3CH.
\end{equation}

We introduce an allocation ratio $\alpha\in[0,1]$, representing the fraction of the expert parameter budget assigned to the gated branch:

\begin{equation}
3CS=\alpha(3CH),
\end{equation}

\begin{equation}
2CR=(1-\alpha)(3CH).
\end{equation}

The corresponding intermediate widths are

\begin{equation}
S=\alpha H,\qquad R=\frac{3}{2}(1-\alpha)H.
\end{equation}

This formulation unifies fully gated, partially gated, and non-gated experts within a fixed parameter budget. By default, we use $\alpha=0.5$, yielding $S=H/2$ and $R=3H/4$.

\section{Experiments}

\subsection{Experimental Setup}

We conduct 100 pretraining runs on two corpora, SmolLM and FineWebEdu, covering MoE models from 30M to 1.03B parameters with a cumulative training budget of approximately 247 billion tokens. The SmolLM experiments comprise 18 main runs and 60 ablation runs, while the FineWebEdu experiments comprise 22 runs across 11 model configurations with three expert granularities. We further conduct four full fine-tuning runs on BillSum and PubMed to evaluate downstream performance.

All models contain 10 Transformer layers, with a dense SwiGLU FFN in the first layer and MoE FFNs in the remaining nine layers. Within each architectural comparison, we match the parameter count, principal dense projection MACs, and training protocol. All pretraining runs use AdamW, BF16 mixed precision, and a context length of 4,096. Each SmolLM model is trained on approximately 1.071 billion tokens with three random seeds (42, 43, and 44), while the FineWebEdu experiments use 100 training tokens per active parameter with Seed 42. We report validation perplexity (PPL) as the primary pretraining metric, using mean and standard deviation across seeds for the SmolLM experiments. Detailed model configurations, training protocols, and per-seed results are provided in the appendices.

\subsection{Main Results}\label{subsec:main-results}

Table~\ref{tab:main-smollm} compares ParGLU with SwiGLU across three model sizes on SmolLM, using the default 50/50 parameter allocation.

\begin{table}[!htbp]
\centering
\ParGLUTableStyle
\caption{Main language modeling results on SmolLM.}
\label{tab:main-smollm}
\begin{tabular}{@{}S[table-format=3.2] S[table-format=2.2] S[table-format=2.3]@{\,}c S[table-format=2.3]@{\,}c c@{}}
\toprule
\multicolumn{2}{c}{Parameters (M)} & \multicolumn{4}{c}{Validation PPL $\downarrow$} & Gain $\uparrow$ \\
\cmidrule(lr){1-2}\cmidrule(lr){3-6}
\multicolumn{1}{c}{Total} & \multicolumn{1}{c}{Active} & \multicolumn{2}{c}{SwiGLU} & \multicolumn{2}{c}{ParGLU} & \\
\cmidrule(lr){1-2}\cmidrule(lr){3-4}\cmidrule(lr){5-6}
30.16 & 15.12 & \textbf{25.790} & \ParGLUSD{0.198} & 25.884 & \ParGLUSD{0.285} & $-0.36\%$ \\
75.48 & 30.35 & 21.123 & \ParGLUSD{0.281} & \textbf{20.703} & \ParGLUSD{0.112} & +1.99\% \\
152.12 & 44.95 & 19.403 & \ParGLUSD{0.135} & \textbf{18.925} & \ParGLUSD{0.171} & +2.46\% \\
\bottomrule
\end{tabular}
\par\smallskip
\begin{minipage}{0.9\linewidth}\footnotesize\raggedright
PPL is reported as mean $\pm$ sample standard deviation across three seeds. Gain denotes the relative reduction in mean PPL.
\end{minipage}
\end{table}

ParGLU consistently outperforms SwiGLU at 75M and 152M, reducing mean PPL by 1.99\% and 2.46\%, respectively, with mixed results at 30M. The relationship between expert width and these improvements is further examined in Section~\ref{sec:backbone-expert-width}.

\subsection{Scaling across Expert Granularities}\label{subsec:scaling-expert-granularities}

We extend our evaluation to 11 model configurations on FineWebEdu, spanning three expert granularities and total parameter counts from 137M to 1.03B.

\begin{table}[!htbp]
\centering
\ParGLUTableStyle
\caption{Scaling results across model sizes and expert granularities on FineWebEdu.}
\label{tab:finewebedu-scaling}
\noindent\textit{(a) Paired validation results}\par\smallskip
\begin{tabular*}{0.94\linewidth}{@{\extracolsep{\fill}}l S[table-format=4.2] S[table-format=3.2] S[table-format=2.2] S[table-format=2.3] S[table-format=2.3] c@{}}
\toprule
\multirow{2}{*}{Granularity} & \multicolumn{2}{c}{Parameters (M)} & \multicolumn{1}{c}{\multirow{2}{*}{Tokens (B)}} & \multicolumn{2}{c}{Validation PPL $\downarrow$} & \multirow{2}{*}{Gain $\uparrow$} \\
\cmidrule(lr){2-3}\cmidrule(lr){5-6}
& \multicolumn{1}{c}{Total} & \multicolumn{1}{c}{Active} & & \multicolumn{1}{c}{SwiGLU} & \multicolumn{1}{c}{ParGLU} & \\
\midrule
Coarse & 137.16 & 38.45 & 3.85 & 21.712 & \textbf{21.354} & +1.65\% \\
Coarse & 246.63 & 59.67 & 5.97 & 18.127 & \textbf{17.960} & +0.92\% \\
Coarse & 395.32 & 84.28 & 8.43 & 16.078 & \textbf{15.980} & +0.61\% \\
Coarse & 648.73 & 119.55 & 11.95 & 14.623 & \textbf{14.580} & +0.30\% \\
Balanced & 253.26 & 48.88 & 4.89 & 18.801 & \textbf{18.534} & +1.42\% \\
Balanced & 401.85 & 70.41 & 7.04 & 16.739 & \textbf{16.612} & +0.76\% \\
Balanced & 642.75 & 95.32 & 9.53 & 15.159 & \textbf{15.094} & +0.42\% \\
Fine & 250.66 & 38.16 & 3.82 & 20.008 & \textbf{19.750} & +1.29\% \\
Fine & 398.97 & 58.57 & 5.86 & 17.046 & \textbf{16.994} & +0.31\% \\
Fine & 634.36 & 80.85 & 8.08 & 15.468 & \textbf{15.415} & +0.34\% \\
Fine & 1029.29 & 123.54 & 12.35 & 13.970 & \textbf{13.849} & +0.86\% \\
\bottomrule
\end{tabular*}
\par\vspace{4pt}
\noindent\textit{(b) Relative PPL reduction across nominal model sizes}\par\smallskip
\begin{tabular*}{0.76\linewidth}{@{\extracolsep{\fill}}lccc@{}}
\toprule
Scale & Coarse & Balanced & Fine \\
\midrule
$\sim$250M & +0.92\% & \textbf{+1.42\%} & +1.29\% \\
$\sim$400M & +0.61\% & \textbf{+0.76\%} & +0.31\% \\
$\sim$650M & +0.30\% & \textbf{+0.42\%} & +0.34\% \\
\bottomrule
\end{tabular*}
\par\smallskip
\begin{minipage}{\linewidth}\footnotesize\raggedright
All experiments use Seed 42 and 100 training tokens per active parameter. Relative PPL reductions are computed within each configuration using unrounded values. Active parameter counts and training budgets differ across expert granularities.
\end{minipage}
\end{table}

ParGLU outperforms SwiGLU in all 11 configurations, with relative PPL reductions ranging from 0.30\% to 1.65\%. Among the three expert granularities, the balanced configurations exhibit the largest relative improvements at all three overlapping nominal model sizes.

\subsection{Gating Ratio and Activation Ablations}\label{subsec:gating-activation-ablations}

We examine different allocations of the expert parameter budget and alternative activation functions in the 75M and 152M MoE models.

\begin{table}[!htbp]
\centering
\ParGLUTableStyle
\caption{Gating ratio and activation ablations on SmolLM.}
\label{tab:gating-activation-ablations}
\begin{tabular*}{\linewidth}{@{\extracolsep{\fill}}l S[table-format=2.3]@{\,}c c S[table-format=2.3]@{\,}c c@{}}
\toprule
\multirow{2}{*}{Architecture} & \multicolumn{3}{c}{75M ($H=256$)} & \multicolumn{3}{c}{152M ($H=608$)} \\
\cmidrule(lr){2-4}\cmidrule(lr){5-7}
& \multicolumn{2}{c}{PPL $\downarrow$} & Gain $\uparrow$ & \multicolumn{2}{c}{PPL $\downarrow$} & Gain $\uparrow$ \\
\midrule
Fully gated: Pure SwiGLU & 21.123 & \ParGLUSD{0.281} & --- & 19.403 & \ParGLUSD{0.135} & --- \\
Fully gated: Pure ReGLU & 21.267 & \ParGLUSD{0.405} & $-0.68\%$ & 19.589 & \ParGLUSD{0.162} & $-0.96\%$ \\
Allocation: Pure ReLU$^2$ (0/100) & 20.732 & \ParGLUSD{0.339} & +1.85\% & 18.965 & \ParGLUSD{0.134} & +2.26\% \\
Allocation: ParGLU 25/75 & 20.793 & \ParGLUSD{0.201} & +1.56\% & \textbf{18.869} & \ParGLUSD{0.114} & +2.75\% \\
Allocation: ParGLU 50/50 & 20.703 & \ParGLUSD{0.112} & +1.99\% & 18.925 & \ParGLUSD{0.171} & +2.46\% \\
Allocation: ParGLU 75/25 & \textbf{20.635} & \ParGLUSD{0.195} & +2.31\% & 18.902 & \ParGLUSD{0.135} & +2.58\% \\
Alternative activation: SwiGLU + GELU 50/50 & 20.872 & \ParGLUSD{0.261} & +1.19\% & 19.037 & \ParGLUSD{0.306} & +1.88\% \\
Alternative activation: SwiGLU + ReLU 50/50 & 21.097 & \ParGLUSD{0.187} & +0.12\% & 19.200 & \ParGLUSD{0.190} & +1.05\% \\
\bottomrule
\end{tabular*}
\par\smallskip
\begin{minipage}{\linewidth}\footnotesize\raggedright
PPL is reported as mean $\pm$ sample standard deviation across three seeds. Ratios denote gated/non-gated parameter budget allocations. Gain is measured against pure SwiGLU.
\end{minipage}
\end{table}

All three partial-gating ratios outperform pure SwiGLU across all 18 paired comparisons. Although pure ReLU$^2$ also improves upon SwiGLU, partially gated configurations achieve the lowest mean PPL at both model sizes. At the 50/50 allocation, ReLU$^2$ also yields lower mean PPL than GELU and ReLU.

\subsection{Backbone and Expert Width Ablation}\label{sec:backbone-expert-width}

We conduct a \((2\times2)\) ablation over backbone dimension \((C\in\{256,384\})\) and expert width \((H\in\{128,256\})\) to examine how model dimensions affect the benefits of partial gating.

\begin{table}[!htbp]
\centering
\ParGLUTableStyle
\caption{Backbone and expert width ablation on SmolLM.}
\label{tab:backbone-expert-width}
\begin{tabular*}{0.92\linewidth}{@{\extracolsep{\fill}}S[table-format=3] S[table-format=3] S[table-format=2.2] S[table-format=2.3]@{\,}c S[table-format=2.3]@{\,}c c@{}}
\toprule
\multicolumn{3}{c}{Model Configuration} & \multicolumn{4}{c}{Validation PPL $\downarrow$} & \multirow{2}{*}{Gain $\uparrow$} \\
\cmidrule(lr){1-3}\cmidrule(lr){4-7}
\multicolumn{1}{c}{C} & \multicolumn{1}{c}{H} & \multicolumn{1}{c}{Total (M)} & \multicolumn{2}{c}{SwiGLU} & \multicolumn{2}{c}{ParGLU} & \\
\cmidrule(lr){1-3}\cmidrule(lr){4-5}\cmidrule(lr){6-7}
256 & 128 & 30.16 & \multicolumn{1}{c}{\textbf{25.790}} & \ParGLUSD{0.198} & 25.884 & \ParGLUSD{0.285} & $-0.36\%$ \\
384 & 128 & 47.61 & 22.637 & \ParGLUSD{0.495} & \multicolumn{1}{c}{\textbf{22.350}} & \ParGLUSD{0.125} & +1.27\% \\
256 & 256 & 48.74 & 23.813 & \ParGLUSD{0.207} & \multicolumn{1}{c}{\textbf{23.195}} & \ParGLUSD{0.248} & +2.60\% \\
384 & 256 & 75.48 & 21.123 & \ParGLUSD{0.281} & \multicolumn{1}{c}{\textbf{20.703}} & \ParGLUSD{0.112} & +1.99\% \\
\bottomrule
\end{tabular*}
\par\smallskip
\begin{minipage}{\linewidth}\footnotesize\raggedright
PPL is reported as mean $\pm$ sample standard deviation across three seeds. The 30.16M and 75.48M results are reused from Table~\ref{tab:main-smollm}.
\end{minipage}
\end{table}

Increasing expert width from 128 to 256 improves the relative performance of ParGLU at both backbone dimensions. ParGLU wins two of six paired comparisons at \((H=128)\), compared with six of six at \((H=256)\).

\subsection{Shared and Routed Expert Ablation}\label{subsec:shared-routed-expert-ablation}

We separately apply partial gating to shared experts, routed experts, and both expert types to examine their contributions to language modeling performance.

\begin{table}[!htbp]
\centering
\ParGLUTableStyle
\caption{Partial gating across shared and routed experts on SmolLM.}
\label{tab:shared-routed-expert-ablation}
\begin{tabular*}{0.90\linewidth}{@{\extracolsep{\fill}}l S[table-format=2.3]@{\,}c c S[table-format=2.3]@{\,}c c@{}}
\toprule
\multirow{2}{*}{Architecture} & \multicolumn{3}{c}{75M ($H=256$)} & \multicolumn{3}{c}{152M ($H=608$)} \\
\cmidrule(lr){2-4}\cmidrule(lr){5-7}
& \multicolumn{2}{c}{PPL $\downarrow$} & Gain $\uparrow$ & \multicolumn{2}{c}{PPL $\downarrow$} & Gain $\uparrow$ \\
\midrule
Fully gated: Pure SwiGLU & 21.123 & \ParGLUSD{0.281} & --- & 19.403 & \ParGLUSD{0.135} & --- \\
Shared-only & 20.898 & \ParGLUSD{0.275} & +1.06\% & 19.257 & \ParGLUSD{0.136} & +0.75\% \\
Routed-only & 20.841 & \ParGLUSD{0.020} & +1.34\% & \multicolumn{1}{c}{\textbf{18.644}} & \ParGLUSD{0.292} & +3.91\% \\
Full ParGLU & \multicolumn{1}{c}{\textbf{20.703}} & \ParGLUSD{0.112} & +1.99\% & 18.925 & \ParGLUSD{0.171} & +2.46\% \\
\bottomrule
\end{tabular*}
\par\smallskip
\begin{minipage}{\linewidth}\footnotesize\raggedright
PPL is reported as mean $\pm$ sample standard deviation across three seeds. Gain is measured against pure SwiGLU. All ParGLU experts use the 50/50 allocation.
\end{minipage}
\end{table}

Both shared-only and routed-only ParGLU improve mean PPL at both model sizes. Full ParGLU performs best at 75M, whereas routed-only ParGLU achieves the lowest PPL across all three seeds at 152M, reducing mean PPL by 3.91\% relative to SwiGLU.

\subsection{Downstream Fine-Tuning}\label{subsec:downstream-fine-tuning}

We independently fine-tune the pretrained 1.03B SwiGLU and ParGLU models on BillSum and PubMed to evaluate downstream summarization performance.

\begin{table}[!htbp]
\centering
\ParGLUTableStyle
\setlength{\tabcolsep}{3pt}
\caption{Downstream full fine-tuning results of the 1.03B models.}
\label{tab:downstream-finetuning}
\sisetup{text-series-to-math}
\begin{tabular*}{\linewidth}{@{\extracolsep{\fill}}>{\centering\arraybackslash}m{1.15in} c S[table-format=2.3,table-model-setup=\bfseries] S[table-format=1.3,table-model-setup=\bfseries] c S[table-format=2.2,table-model-setup=\bfseries] S[table-format=2.2,table-model-setup=\bfseries] S[table-format=2.2,table-model-setup=\bfseries]@{}}
\toprule
\multirow{2}{*}{Test Set} & \multirow{2}{*}{Model} & \multicolumn{3}{c}{Perplexity $\downarrow$} & \multicolumn{3}{c}{Final ROUGE $\uparrow$} \\
\cmidrule(lr){3-5}\cmidrule(lr){6-8}
& & \multicolumn{1}{c}{Base} & \multicolumn{1}{c}{SFT} & Gain $\uparrow$ & \multicolumn{1}{c}{R-1} & \multicolumn{1}{c}{R-2} & \multicolumn{1}{c}{R-L} \\
\midrule
\multirow{2}{*}{\parbox[c]{1.15in}{\centering BillSum US}} & SwiGLU & 7.877 & 2.825 & --- & 48.93 & 30.64 & 37.47 \\
& ParGLU & \bfseries 7.434 & \bfseries 2.811 & +0.50\% & \bfseries 49.61 & \bfseries 30.66 & \bfseries 38.07 \\
\addlinespace[4pt]
\multirow{2}{*}{\parbox[c]{1.15in}{\centering BillSum CA}} & SwiGLU & 6.831 & 6.419 & --- & 34.89 & 15.14 & 22.15 \\
& ParGLU & \bfseries 6.693 & \bfseries 6.290 & +2.01\% & \bfseries 39.14 & \bfseries 16.72 & \bfseries 24.40 \\
\addlinespace[4pt]
\multirow{2}{*}{\parbox[c]{1.15in}{\centering PubMed}} & SwiGLU & \bfseries 12.526 & 5.488 & --- & \bfseries 38.07 & \bfseries 14.51 & 23.58 \\
& ParGLU & 12.855 & \bfseries 5.455 & +0.61\% & 37.74 & 14.42 & \bfseries 23.89 \\
\bottomrule
\end{tabular*}
\par\smallskip
\begin{minipage}{\linewidth}\footnotesize\raggedright
PPL is evaluated on the full test sets before and after fine-tuning. ROUGE scores are reported on a 0--100 scale using 256 sampled examples per test set, with greedy decoding and a maximum of 256 generated tokens. All fine-tuning runs use Seed 42. Gain denotes the relative PPL reduction after fine-tuning compared with the fine-tuned SwiGLU baseline.
\end{minipage}
\end{table}

ParGLU achieves lower PPL on all three test sets and outperforms SwiGLU on seven of nine ROUGE comparisons. Notably, on PubMed, ParGLU starts with a higher pretrained PPL but achieves a lower PPL after fine-tuning.

\subsection{Training Throughput}\label{subsec:training-throughput}

We compare the training throughput of SwiGLU and ParGLU across all 11 FineWebEdu configurations using the throughput recorded during pretraining.

\begin{table}[!htbp]
\centering
\ParGLUTableStyle
\caption{Training throughput across model sizes and expert granularities on FineWebEdu.}
\label{tab:training-throughput}
\begin{tabular*}{0.90\linewidth}{@{\extracolsep{\fill}}l S[table-format=3.2] S[table-format=3.2] S[table-format=3.2]@{}}
\toprule
\multirow{2}{*}{Configuration} & \multicolumn{2}{c}{Throughput (K tok/s)} & \multicolumn{1}{c}{\multirow{2}{*}{Ratio (\%)}} \\
\cmidrule(lr){2-3}
& \multicolumn{1}{c}{SwiGLU} & \multicolumn{1}{c}{ParGLU} & \\
\midrule
Coarse 137M & \textbf{235.34} & 228.83 & 97.23 \\
Coarse 250M & \textbf{149.55} & 131.24 & 87.76 \\
Coarse 395M & \textbf{118.51} & 106.31 & 89.71 \\
Coarse 649M & \textbf{80.55} & 80.38 & 99.79 \\
Balanced 253M & 186.16 & \textbf{186.41} & 100.13 \\
Balanced 402M & \textbf{142.27} & 139.54 & 98.09 \\
Balanced 643M & \textbf{109.99} & 109.71 & 99.74 \\
Fine 251M & \textbf{183.22} & 183.15 & 99.96 \\
Fine 399M & 122.00 & \textbf{126.16} & 103.41 \\
Fine 634M & 114.29 & \textbf{114.81} & 100.45 \\
Fine 1.03B & \textbf{84.35} & 81.59 & 96.73 \\
\bottomrule
\end{tabular*}
\par\smallskip
\begin{minipage}{0.90\linewidth}\footnotesize\raggedright
Throughput is reported as the median of nine mean throughput measurements over consecutive 1,000-step intervals from steps 1,001 to 10,000. Ratio denotes ParGLU throughput relative to SwiGLU.
\end{minipage}
\end{table}

ParGLU achieves comparable training throughput to SwiGLU in most configurations, retaining at least 96.7\% of the baseline throughput in nine of the eleven comparisons. The 1.03B model retains 96.73\% of SwiGLU throughput while achieving lower validation perplexity.

\section{Analysis}\label{sec:analysis}

\subsection{Token Frequency Analysis}\label{subsec:token-frequency-analysis}

To examine how the benefits of partial gating vary across token frequencies, we partition validation targets into four groups based on their frequencies in the training corpus, from the least frequent (Q1) to the most frequent (Q4).

\begin{table}[tb]
\centering
\footnotesize
\setlength{\tabcolsep}{4pt}
\renewcommand{\arraystretch}{1.12}
\caption{Token-frequency and predictive-entropy NLL differences.}
\label{tab:token-frequency-groups}
\label{tab:entropy-quartiles}
\begin{tabular}{@{}lrrrrrrrr@{}}
\toprule
& \multicolumn{4}{c}{Token Frequency} & \multicolumn{4}{c}{Predictive Entropy} \\
\cmidrule(lr){2-5}\cmidrule(lr){6-9}
Scale & Q1 & Q2 & Q3 & Q4 & Q1 & Q2 & Q3 & Q4 \\
\midrule
\rowcolor{GrpBg}
\multicolumn{9}{l}{\textbf{SmolLM}} \\
30M  & $+0.0051$ & $+0.0025$ & $+0.0029$ & $+0.0039$ & \multicolumn{4}{c}{} \\
75M  & $\boldsymbol{-0.0283}$ & $-0.0251$ & $-0.0189$ & $-0.0079$ & \multicolumn{4}{c}{\textit{Not evaluated}} \\
152M & $\boldsymbol{-0.0356}$ & $-0.0299$ & $-0.0221$ & $-0.0121$ & \multicolumn{4}{c}{} \\
\rowcolor{GrpBg}
\multicolumn{9}{l}{\textbf{FineWebEdu: Coarse}} \\
137M & $-0.0222$ & $\boldsymbol{-0.0232}$ & $-0.0164$ & $-0.0050$ & $-0.0009$ & $-0.0134$ & $-0.0208$ & $\boldsymbol{-0.0315}$ \\
250M & $-0.0136$ & $\boldsymbol{-0.0144}$ & $-0.0086$ & $-0.0008$ & $+0.0022$ & $-0.0039$ & $-0.0124$ & $\boldsymbol{-0.0232}$ \\
395M & $\boldsymbol{-0.0109}$ & $-0.0061$ & $-0.0028$ & $-0.0045$ & $+0.0031$ & $-0.0013$ & $-0.0091$ & $\boldsymbol{-0.0169}$ \\
649M & $-0.0030$ & $-0.0033$ & $\boldsymbol{-0.0042}$ & $-0.0014$ & $+0.0024$ & $-0.0008$ & $-0.0039$ & $\boldsymbol{-0.0098}$ \\
\rowcolor{GrpBg}
\multicolumn{9}{l}{\textbf{FineWebEdu: Balanced}} \\
253M & $\boldsymbol{-0.0229}$ & $-0.0176$ & $-0.0101$ & $-0.0067$ & $+0.0003$ & $-0.0085$ & $-0.0179$ & $\boldsymbol{-0.0314}$ \\
402M & $\boldsymbol{-0.0136}$ & $-0.0126$ & $-0.0041$ & $-0.0002$ & $+0.0022$ & $-0.0032$ & $-0.0113$ & $\boldsymbol{-0.0180}$ \\
643M & $-0.0070$ & $\boldsymbol{-0.0084}$ & $-0.0047$ & $+0.0029$ & $+0.0025$ & $-0.0007$ & $-0.0055$ & $\boldsymbol{-0.0134}$ \\
\rowcolor{GrpBg}
\multicolumn{9}{l}{\textbf{FineWebEdu: Fine}} \\
251M & $\boldsymbol{-0.0178}$ & $-0.0174$ & $-0.0097$ & $-0.0073$ & $+0.0011$ & $-0.0085$ & $-0.0169$ & $\boldsymbol{-0.0276}$ \\
399M & $-0.0031$ & $-0.0018$ & $\boldsymbol{-0.0054}$ & $-0.0022$ & $+0.0037$ & $-0.0005$ & $-0.0038$ & $\boldsymbol{-0.0117}$ \\
634M & $\boldsymbol{-0.0073}$ & $-0.0018$ & $-0.0036$ & $-0.0012$ & $+0.0029$ & $-0.0010$ & $-0.0048$ & $\boldsymbol{-0.0110}$ \\
1.03B & $\boldsymbol{-0.0151}$ & $-0.0107$ & $-0.0059$ & $-0.0031$ & $+0.0008$ & $-0.0036$ & $-0.0118$ & $\boldsymbol{-0.0202}$ \\
\bottomrule
\end{tabular}
\par\smallskip
\parbox{\linewidth}{\footnotesize Values denote $\Delta\mathrm{NLL}$ (ParGLU $-$ SwiGLU). Negative values indicate improvements. SmolLM results are averaged across three seeds; FineWebEdu results use Seed 42.}
\end{table}

ParGLU exhibits stronger improvements on low-frequency tokens. Across all thirteen configurations with overall improvements in Table~\ref{tab:token-frequency-groups}, the NLL reduction is larger in Q1 than in Q4. On FineWebEdu, ParGLU reduces NLL in 43 of the 44 frequency-group comparisons.

\subsection{Predictive Entropy Analysis}\label{subsec:predictive-entropy-analysis}

We further examine whether the benefits of partial gating vary with predictive uncertainty. Validation targets are divided into four equal-sized groups according to the predictive entropy of the corresponding SwiGLU baseline, from the lowest (Q1) to the highest (Q4), as shown in Table~\ref{tab:entropy-quartiles}.

Across all eleven configurations, the NLL advantage of ParGLU increases monotonically with predictive entropy. ParGLU improves NLL in all 33 comparisons across Q2--Q4, with the largest gains consistently occurring in Q4. The joint frequency--entropy analysis further reveals that these benefits extend to high-frequency tokens: at 1.03B, the highest-frequency and highest-entropy group achieves an NLL reduction of 0.03618. These results highlight the effectiveness of partial gating on uncertain predictions across token frequencies.

\subsection{Native Activation Sparsity}\label{subsec:native-activation-sparsity}

The squared-ReLU branch of ParGLU naturally produces exact zero activations. We quantify this native sparsity in the pretrained 1.03B model over the full FineWebEdu validation set, without activation pruning or sparsity regularization.

\begin{table}[!htbp]
\centering
\ParGLUTableStyle
\caption{Native activation sparsity and theoretical MAC reduction in the 1.03B ParGLU model.}
\label{tab:native-activation-sparsity}
\begin{tabular}{@{}lr@{}}
\toprule
Metric & Value \\
\midrule
\multicolumn{2}{@{}l}{\textit{Activation sparsity}}\\
SwiGLU branch & $\approx 0.00\%$ \\
Squared-ReLU branch (shared experts) & 83.87\% \\
Squared-ReLU branch (routed experts) & 86.37\% \\
Squared-ReLU branch (overall) & \textbf{85.75\%} \\
\addlinespace[2pt]
\multicolumn{2}{@{}l}{\textit{Theoretical MAC reduction}}\\
Expert projections & \textbf{21.44\%} \\
Full-model principal linear projections & 9.67\% \\
\bottomrule
\end{tabular}
\par\smallskip
\begin{minipage}{0.90\linewidth}\footnotesize\raggedright
Statistics are collected across 193,314,816 expert invocations in nine MoE layers. MAC savings correspond to skipping zero-valued intermediate contributions in the down-projections.
\end{minipage}
\end{table}

As shown in Table~\ref{tab:native-activation-sparsity}, 85.75\% of squared-ReLU activations are exactly zero, with substantial sparsity in both shared and routed experts. Skipping the corresponding down-projection operations reduces theoretical expert projection MACs by 21.44\%, equivalent to a 9.67\% reduction across the principal linear projections of the entire model. This computational potential arises directly from the architecture, without introducing additional sparsity constraints.

The derivation of theoretical MAC savings is provided in Appendix~\ref{app:additional-sparsity}.

\subsection{Quality--Computation Trade-off}\label{subsec:quality-computation-tradeoff}

To evaluate the quality--computation trade-off of native sparsity, we compare ParGLU with SwiGLU under varying levels of post-hoc magnitude pruning. For SwiGLU, only the top-$K$ intermediate activations by magnitude are retained before the down-projection.

\begin{table}[!htbp]
\centering
\ParGLUTableStyle
\caption{Quality--computation trade-off of magnitude-pruned SwiGLU and native ParGLU at 1.03B.}
\label{tab:quality-computation-tradeoff}
\begin{tabular}{@{}lrrr@{}}
\toprule
Architecture & Retained $K$ & Expert MAC Reduction & PPL \\
\midrule
SwiGLU & 576 & 0.00\% & 13.970 \\
SwiGLU & 461 & 6.66\% & 13.971 \\
SwiGLU & 346 & 13.31\% & 13.988 \\
SwiGLU & 288 & 16.67\% & 14.020 \\
SwiGLU & 204 & 21.53\% & 14.175 \\
SwiGLU & 173 & 23.32\% & 14.319 \\
SwiGLU & 144 & 25.00\% & 14.554 \\
SwiGLU & 115 & 26.68\% & 14.963 \\
Native ParGLU & --- & 21.44\% & \textbf{13.849} \\
\bottomrule
\end{tabular}
\par\smallskip
\begin{minipage}{0.90\linewidth}\footnotesize\raggedright
MAC reductions are measured relative to unpruned SwiGLU. ParGLU uses native activation sparsity without magnitude pruning. Detailed evaluation settings are provided in Appendix E.
\end{minipage}
\end{table}

Native ParGLU reduces expert projection MACs by 21.44\% while achieving a lower PPL than unpruned SwiGLU. At nearly identical theoretical computation costs, ParGLU achieves a PPL of 13.849, compared with 14.175 for magnitude-pruned SwiGLU, corresponding to a relative reduction of 2.30\%.

\begin{table}[!htbp]
\centering
\ParGLUTableStyle
\caption{Effective projection widths under dense and sparse computation.}
\label{tab:effective-projection-widths}
\begin{tabular}{@{}lrrrr@{}}
\toprule
Architecture & Up/Gate & Down & Total & PPL \\
\midrule
Dense SwiGLU & 1152 & 576 & 1728 & 13.970 \\
Dense ParGLU & 1008 & 720 & 1728 & \textbf{13.849} \\
Pruned SwiGLU ($K=204$) & 1152 & 204 & 1356 & 14.175 \\
Native ParGLU & 1008 & 349.57 & 1357.57 & \textbf{13.849} \\
\bottomrule
\end{tabular}
\par\smallskip
\begin{minipage}{0.90\linewidth}\footnotesize\raggedright
Effective projection widths multiplied by the backbone dimension give the corresponding expert projection MACs. Dense and native ParGLU refer to the same pretrained model.
\end{minipage}
\end{table}
\FloatBarrier

This advantage is accompanied by a redistribution of projection capacity. Under nearly identical effective MAC budgets, ParGLU retains 71.4\% more active down-projection channels than magnitude-pruned SwiGLU, while allocating fewer operations to the up-projections. Partial gating therefore provides a more favorable quality--computation trade-off by combining wider intermediate representations with native activation sparsity.

\section{Conclusion}

MoE architectures have benefited from distributing capacity across experts. ParGLU extends this principle to the interior of each expert, demonstrating that multiplicative gating is valuable without requiring the entire parameter budget. Through a simple combination of SwiGLU and squared-ReLU, ParGLU improves language modeling while preserving parameter count, dense projection computation, and comparable training throughput.

Our findings establish intra-expert parameter allocation as an effective design dimension for MoE Transformers. By combining multiplicative gating, wider intermediate representations, and native activation sparsity, partial gating offers a practical path toward more capable and computationally efficient experts.

\section*{Limitations}

\textbf{Model scale.} Our experiments extend to 1.03B parameters. Larger models and longer training horizons offer further opportunities to explore the scaling behavior of partial gating.

\textbf{Allocation ratio.} The optimal gating budget may vary with model dimensions, expert granularity, and training scale.

\textbf{Expert specialization.} The different outcomes for shared and routed experts motivate further exploration of heterogeneous gating budgets within MoE layers.

\textbf{Architectural generalization.} This work focuses on decoder-only MoE Transformers. Partial gating can be explored in dense Transformers and alternative architectures.

\textbf{Downstream applications.} Our fine-tuning experiments focus on document summarization. Broader applications include reasoning, code generation, and instruction following.

\textbf{Computational efficiency.} Translating native activation sparsity into practical inference acceleration requires dedicated sparsity-aware implementations.

\bibliographystyle{plainnat}
\bibliography{references}

@InProceedings{Vaswani2017,
  author    = {Vaswani, Ashish and Shazeer, Noam and Parmar, Niki and Uszkoreit, Jakob and Jones, Llion and Gomez, Aidan N. and Kaiser, {\L}ukasz and Polosukhin, Illia},
  title     = {Attention Is All You Need},
  booktitle = {Advances in Neural Information Processing Systems},
  volume    = {30},
  year      = {2017},
  url       = {https://papers.nips.cc/paper_files/paper/2017/hash/3f5ee243547dee91fbd053c1c4a845aa-Abstract.html}
}

@Misc{HendrycksGimpel2016,
  author       = {Hendrycks, Dan and Gimpel, Kevin},
  title        = {Gaussian Error Linear Units (GELUs)},
  howpublished = {arXiv preprint arXiv:1606.08415},
  year         = {2016},
  url          = {https://arxiv.org/abs/1606.08415}
}

@InProceedings{Dauphin2017,
  author    = {Dauphin, Yann N. and Fan, Angela and Auli, Michael and Grangier, David},
  title     = {Language Modeling with Gated Convolutional Networks},
  booktitle = {Proceedings of the 34th International Conference on Machine Learning},
  pages     = {933--941},
  year      = {2017},
  editor    = {Precup, Doina and Teh, Yee Whye},
  volume    = {70},
  series    = {Proceedings of Machine Learning Research},
  publisher = {PMLR},
  url       = {https://proceedings.mlr.press/v70/dauphin17a.html}
}

@Misc{Shazeer2020,
  author       = {Shazeer, Noam},
  title        = {GLU Variants Improve Transformer},
  howpublished = {arXiv preprint arXiv:2002.05202},
  year         = {2020},
  url          = {https://arxiv.org/abs/2002.05202}
}

@InProceedings{So2021,
  author    = {So, David R. and Ma\'{n}ke, Wojciech and Liu, Hanxiao and Dai, Zihang and Shazeer, Noam and Le, Quoc V.},
  title     = {Searching for Efficient Transformers for Language Modeling},
  booktitle = {Advances in Neural Information Processing Systems},
  volume    = {34},
  year      = {2021},
  url       = {https://papers.nips.cc/paper_files/paper/2021/hash/2f3c6a4cd8af177f6456e7e51a916ff3-Abstract.html}
}

@Misc{Arefin2026,
  author       = {Arefin, Md Rifat and Heidari, Farzaneh and Rish, Irina and Rabusseau, Guillaume},
  title        = {{CP-MLPs: A Tensor-Rank Theory of Tied and Untied MLP Blocks}},
  howpublished = {arXiv preprint arXiv:2611.13647},
  year         = {2026},
  url          = {https://arxiv.org/abs/2611.13647}
}

@InProceedings{Dai2024DeepSeekMoE,
  author    = {Dai, Damai and Deng, Chengqi and Zhao, Chenggang and Xu, R. X. and Gao, Huazuo and Chen, Deli and Li, Jiashi and Zeng, Wangding and Yu, Xingkai and Wu, Y. and Xie, Zhenda and Li, Y. K. and Huang, Panpan and Luo, Fuli and Ruan, Chong and Sui, Zhifang and Liang, Wenfeng},
  title     = {{DeepSeekMoE: Towards Ultimate Expert Specialization in Mixture-of-Experts Language Models}},
  booktitle = {Proceedings of the 62nd Annual Meeting of the Association for Computational Linguistics (Volume 1: Long Papers)},
  pages     = {1280--1297},
  year      = {2024},
  month     = {August},
  address   = {Bangkok, Thailand},
  publisher = {Association for Computational Linguistics},
  url       = {https://aclanthology.org/2024.acl-long.70/}
}

@InProceedings{Ludziejewski2024Scaling,
  author    = {Ludziejewski, Jan and Krajewski, Jakub and Adamczewski, Kamil and Pi{\'o}ro, Maciej and Krutul, Micha{\l} and Antoniak, Szymon and Ciebiera, Kamil and Kr{\'o}l, Krystian and Odrzyg{\'o}{\'z}d{\'z}, Tomasz and Sankowski, Piotr and Cygan, Marek and Jaszczur, Sebastian},
  title     = {Scaling Laws for Fine-Grained Mixture of Experts},
  booktitle = {Proceedings of the 41st International Conference on Machine Learning},
  pages     = {33270--33288},
  year      = {2024},
  volume    = {235},
  series    = {Proceedings of Machine Learning Research},
  publisher = {PMLR},
  url       = {https://proceedings.mlr.press/v235/ludziejewski24a.html}
}

@InProceedings{pmlr-v267-ludziejewski25a,
  author    = {Ludziejewski, Jan and Pi{\'o}ro, Maciej and Krajewski, Jakub and Stefaniak, Maciej and Krutul, Micha{\l} and Ma{\l}a\'{s}nicki, Jan and Cygan, Marek and Sankowski, Piotr and Adamczewski, Kamil and Mi{\l}o\'{s}, Piotr and Jaszczur, Sebastian},
  title     = {Joint {M}o{E} Scaling Laws: Mixture of Experts Can Be Memory Efficient},
  booktitle = {Proceedings of the 42nd International Conference on Machine Learning},
  pages     = {41056--41073},
  year      = {2025},
  volume    = {267},
  series    = {Proceedings of Machine Learning Research},
  publisher = {PMLR},
  url       = {https://proceedings.mlr.press/v267/ludziejewski25a.html}
}

@Misc{Li2026SlicingDicing,
  author       = {Li, Margaret and Kudugunta, Sneha and Rothermel, Danielle and Zettlemoyer, Luke},
  title        = {Slicing and Dicing: Configuring Optimal Mixtures of Experts},
  howpublished = {arXiv preprint arXiv:2605.11689},
  year         = {2026},
  url          = {https://arxiv.org/abs/2605.11689}
}

\section*{Appendices}

\appendix

\setcounter{table}{0}
\renewcommand{\thetable}{\Alph{section}\arabic{table}}
\renewcommand{\theHtable}{\Alph{section}\arabic{table}}
\section{Experimental Details}

\subsection{Model architectures}

All models contain ten Transformer layers. The first layer uses a dense SwiGLU FFN, and the remaining nine layers use MoE FFNs with three activated routed experts and one shared expert per token. The SmolLM models use 20 routed experts per MoE layer. The FineWebEdu configurations vary backbone width, expert width, and the number of routed experts. Within each paired comparison, SwiGLU and ParGLU share the same backbone and routing configuration. Expert and router weights are initialized from a normal distribution with mean zero and standard deviation 0.02.

\begingroup
\scriptsize
\setlength{\tabcolsep}{2pt}
\begin{longtable}{@{}p{1.55in}rrrrrrr@{}}
\caption{Model configurations.}\label{tab:model-configurations}\\
\toprule
Configuration & Total (M) & Active (M) & $C$ & $H$ & Routed $E$ & Dense $H$ & Targets (B) \\
\midrule
\endfirsthead
\toprule
Configuration & Total (M) & Active (M) & $C$ & $H$ & Routed $E$ & Dense $H$ & Targets (B) \\
\midrule
\endhead
\midrule
\multicolumn{8}{r}{Continued on next page}\\
\endfoot
\bottomrule
\endlastfoot
SmolLM 30M & 30.16 & --- & 256 & 128 & 20 & 672 & --- \\
SmolLM C384\_H128 & 47.61 & --- & 384 & 128 & 20 & 1024 & --- \\
SmolLM C256\_H256 & 48.74 & --- & 256 & 256 & 20 & 672 & --- \\
SmolLM 75M & 75.48 & --- & 384 & 256 & 20 & 1024 & --- \\
SmolLM 152M & 152.12 & --- & 384 & 608 & 20 & 1024 & --- \\
FineWebEdu Coarse 137M & 137.157 & 38.454 & --- & --- & 20 & --- & 3.845408 \\
FineWebEdu Coarse 250M & 246.628 & 59.672 & 448 & 736 & 24 & 1184 & 5.967226 \\
FineWebEdu Coarse 395M & 395.321 & 84.281 & 576 & 800 & 28 & 1536 & 8.428090 \\
FineWebEdu Coarse 649M & 648.731 & 119.548 & 704 & 960 & 32 & 1888 & 11.954835 \\
FineWebEdu Balanced 253M & 253.257 & 48.883 & 448 & 512 & 36 & 1184 & 4.888262 \\
FineWebEdu Balanced 402M & 401.853 & 70.409 & 576 & 576 & 40 & 1536 & 7.040851 \\
FineWebEdu Balanced 643M & 642.750 & 95.319 & 704 & 640 & 48 & 1888 & 9.531949 \\
FineWebEdu Fine 251M & 250.660 & 38.158 & 448 & 288 & 64 & 1184 & 3.815750 \\
FineWebEdu Fine 399M & 398.970 & 58.568 & 576 & 384 & 60 & 1536 & 5.856826 \\
FineWebEdu Fine 634M & 634.361 & 80.848 & 704 & 448 & 68 & 1888 & 8.084806 \\
FineWebEdu Fine 1.03B & 1029.290 & 123.541 & 896 & 576 & 68 & 2400 & 12.354138 \\
\end{longtable}
\endgroup

The FineWebEdu models use a head dimension of 64 and tied input embeddings and output weights. Routed experts are grouped in sets of four. Routing selects the two groups with the largest sums of their two highest scores, then selects three experts from the selected groups. The selected experts are weighted by normalized unbiased sigmoid scores. The dynamic routing bias update coefficient is 0.001. Active parameters include the shared backbone, router, dense first layer, and four activated experts per MoE layer.

\subsection{Training and evaluation}

The SmolLM experiments use the ecreeth/1b-smollm-corpus dataset and seeds 42, 43, and 44. The FineWebEdu experiments use seed 42 and a training budget of 100 prediction targets per active parameter. All evaluations use the final checkpoint and token-weighted negative log-likelihood (NLL); perplexity (PPL) is its exponential. Mean PPL across seeds is the arithmetic mean of individual seed PPLs.

\begin{table}[!htbp]
\centering
\ParGLUTableStyle
\caption{Pretraining protocols.}
\begin{tabular}{@{}lrr@{}}
\toprule
Setting & SmolLM & FineWebEdu \\
\midrule
Vocabulary size & 32,768 & 32,000 \\
Context length & 4,096 & 4,096 \\
Global batch (tokens) & 131,072 & 131,072 \\
Optimizer & AdamW & AdamW \\
Betas & (0.9, 0.95) & (0.9, 0.95) \\
Epsilon & 1e-8 & 1e-8 \\
Weight decay & 0.1 & 0.1 \\
Gradient clipping & 1.0 & 1.0 \\
Precision & BF16 autocast / FP32 parameters & BF16 autocast / FP32 parameters \\
Initialization & Normal(0, 0.02) & Normal(0, 0.02) \\
Peak LR & 6e-4 & Configuration-dependent \\
Scheduler & WSD80 & WSD \\
Training targets & 1,071,139,945 & 100 $\times$ active parameters \\
Validation targets & 5,368,638 & 5,368,545 \\
Seeds & 42, 43, 44 & 42 \\
\bottomrule
\end{tabular}
\end{table}

\subsection{Parameter allocation and experiment count}

For a gated budget fraction $\alpha$, branch widths are $S=\alpha H$ and $R=3(1-\alpha)H/2$. Mixed-activation ablations retain the same branch budget while replacing squared ReLU with GELU or ReLU. Shared-only and routed-only variants apply ParGLU to the corresponding expert type and retain SwiGLU in the other. The 100 distinct pretraining runs comprise 18 SmolLM main runs, 36 additional ratio/activation runs, 12 additional width runs, 12 additional expert-type runs, and 22 FineWebEdu scaling runs. Reused baselines are counted once.

\setcounter{table}{0}
\renewcommand{\thetable}{\Alph{section}\arabic{table}}
\renewcommand{\theHtable}{\Alph{section}\arabic{table}}
\section{Full Pretraining Results}

The tables report final validation results for all 100 distinct pretraining runs. Each NLL and PPL entry corresponds to a single model and seed. Lower values are better.

\begingroup
\scriptsize
\setlength{\tabcolsep}{1.5pt}
\begin{longtable}{@{}>{\raggedright\arraybackslash}p{1.55in}>{\raggedright\arraybackslash}p{1.20in}rrrrrr@{}}
\caption{Per-seed validation results for SmolLM.}\label{tab:appendix-b-smollm}\\
\toprule
Experiment / configuration & Architecture & \multicolumn{6}{c}{Validation by seed}\\
\cmidrule(lr){3-8}
& & NLL 42 & PPL 42 & NLL 43 & PPL 43 & NLL 44 & PPL 44\\
\midrule
\endfirsthead
\toprule
Experiment / configuration & Architecture & \multicolumn{6}{c}{Validation by seed}\\
\cmidrule(lr){3-8}
& & NLL 42 & PPL 42 & NLL 43 & PPL 43 & NLL 44 & PPL 44\\
\midrule
\endhead
\midrule
\multicolumn{8}{r}{Continued on next page}\\
\endfoot
\bottomrule
\endlastfoot
Main, 30M & ParGLU & 3.253688 & 25.885629 & 3.242496 & 25.597532 & 3.264528 & 26.167770\\
Main, 30M & SwiGLU & 3.243776 & 25.630310 & 3.247576 & 25.727908 & 3.258530 & 26.011283\\
Main, 75M & ParGLU & 3.023994 & 20.573298 & 3.033163 & 20.762801 & 3.033593 & 20.771722\\
Main, 75M & SwiGLU & 3.064122 & 21.415660 & 3.049117 & 21.096698 & 3.037634 & 20.855844\\
Main, 152M & ParGLU & 2.930243 & 18.732182 & 2.947331 & 19.055029 & 2.943850 & 18.988805\\
Main, 152M & SwiGLU & 2.957356 & 19.247021 & 2.969614 & 19.484400 & 2.969216 & 19.476647\\
Gating / activation, 75M & ParGLU (25/75) & 3.038758 & 20.879305 & 3.041550 & 20.937679 & 3.023516 & 20.563469\\
Gating / activation, 75M & ParGLU (75/25) & 3.018516 & 20.460914 & 3.037163 & 20.846010 & 3.025136 & 20.596802\\
Gating / activation, 75M & Pure ReGLU & 3.066905 & 21.475334 & 3.069218 & 21.525063 & 3.034966 & 20.800268\\
Gating / activation, 75M & Pure ReLU$^2$ & 3.048075 & 21.074740 & 3.015356 & 20.396359 & 3.031403 & 20.726288\\
Gating / activation, 75M & SwiGLU + GELU (50/50) & 3.041322 & 20.932889 & 3.024612 & 20.586010 & 3.049171 & 21.097849\\
Gating / activation, 75M & SwiGLU + ReLU (50/50) & 3.058961 & 21.305401 & 3.041859 & 20.944140 & 3.046470 & 21.040947\\
Gating / activation, 152M & ParGLU (25/75) & 2.937430 & 18.867288 & 2.931528 & 18.756261 & 2.943581 & 18.983707\\
Gating / activation, 152M & ParGLU (75/25) & 2.933105 & 18.785876 & 2.937569 & 18.869921 & 2.947025 & 19.049203\\
Gating / activation, 152M & Pure ReGLU & 2.969815 & 19.488309 & 2.984501 & 19.776628 & 2.970592 & 19.503471\\
Gating / activation, 152M & Pure ReLU$^2$ & 2.940556 & 18.926363 & 2.936777 & 18.854976 & 2.950428 & 19.114123\\
Gating / activation, 152M & SwiGLU + GELU (50/50) & 2.964753 & 19.389913 & 2.937709 & 18.872567 & 2.936460 & 18.849000\\
Gating / activation, 152M & SwiGLU + ReLU (50/50) & 2.963618 & 19.367922 & 2.944135 & 18.994227 & 2.956841 & 19.237114\\
Backbone / expert width, C384\_H128 & ParGLU & 3.113106 & 22.490801 & 3.102560 & 22.254840 & 3.104738 & 22.303369\\
Backbone / expert width, C384\_H128 & SwiGLU & 3.112823 & 22.484437 & 3.143725 & 23.190086 & 3.101691 & 22.235515\\
Backbone / expert width, C256\_H256 & ParGLU & 3.142031 & 23.150837 & 3.134300 & 22.972543 & 3.155383 & 23.462028\\
Backbone / expert width, C256\_H256 & SwiGLU & 3.160740 & 23.588055 & 3.177895 & 23.996191 & 3.172000 & 23.855147\\
Shared / routed experts, 75M & Routed-only ParGLU & 3.036219 & 20.826348 & 3.037973 & 20.862921 & 3.036524 & 20.832697\\
Shared / routed experts, 75M & Shared-only ParGLU & 3.047669 & 21.066185 & 3.046777 & 21.047400 & 3.024383 & 20.581295\\
Shared / routed experts, 152M & Routed-only ParGLU & 2.916536 & 18.477179 & 2.943447 & 18.981155 & 2.916317 & 18.473121\\
Shared / routed experts, 152M & Shared-only ParGLU & 2.959224 & 19.283006 & 2.964100 & 19.377264 & 2.950209 & 19.109951\\
\end{longtable}
\endgroup

\subsection{FineWebEdu scaling}

All models use seed 42. The values below are the formal pretraining validation results, separate from the sparsity evaluation in Appendix E.

\begin{table}[!htbp]
\centering
\footnotesize
\renewcommand{\arraystretch}{1.15}
\setlength{\tabcolsep}{3pt}
\caption{FineWebEdu scaling pretraining results.}
\label{tab:appendix-b-finewebedu}
\begin{tabular}{@{}lrrrrr@{}}
\toprule
Configuration & SwiGLU NLL & ParGLU NLL & $\Delta$NLL & SwiGLU PPL & ParGLU PPL \\
\midrule
Coarse 137M & 3.077876 & 3.061235 & -0.016641 & 21.712246 & 21.353922 \\
Coarse 250M & 2.897426 & 2.888159 & -0.009267 & 18.127421 & 17.960215 \\
Coarse 395M & 2.777447 & 2.771367 & -0.006080 & 16.077922 & 15.980460 \\
Coarse 649M & 2.682579 & 2.679624 & -0.002955 & 14.622763 & 14.579614 \\
Balanced 253M & 2.933932 & 2.919619 & -0.014313 & 18.801412 & 18.534224 \\
Balanced 402M & 2.817737 & 2.810138 & -0.007599 & 16.738927 & 16.612215 \\
Balanced 643M & 2.718578 & 2.714323 & -0.004255 & 15.158749 & 15.094387 \\
Fine 251M & 2.996136 & 2.983129 & -0.013007 & 20.008076 & 19.749507 \\
Fine 399M & 2.835930 & 2.832845 & -0.003085 & 17.046250 & 16.993747 \\
Fine 634M & 2.738762 & 2.735310 & -0.003452 & 15.467819 & 15.414518 \\
Fine 1.03B & 2.636878 & 2.628218 & -0.008660 & 13.969527 & 13.849070 \\
\bottomrule
\end{tabular}
\end{table}

\setcounter{table}{0}
\renewcommand{\thetable}{\Alph{section}\arabic{table}}
\renewcommand{\theHtable}{\Alph{section}\arabic{table}}
\section{Additional Token-Level Analysis}\label{app:additional-token-analysis}

\subsection{Token-frequency grouping}

For SmolLM, token frequencies are estimated from 16 uniformly sampled contiguous training segments of 1,048,576 tokens each, giving 16,777,216 tokens in total. Frequency boundaries are fixed across models and seeds. The validation set is not used to estimate these boundaries. Group NLL is the mean negative log-likelihood over all validation targets assigned to that group.

\begin{table}[!htbp]
\centering
\ParGLUTableStyle
\caption{SmolLM token-frequency groups and validation target counts.}
\label{tab:appendix-c-frequency-boundaries}
\begin{tabular}{@{}lrr@{}}
\toprule
Group & Training frequency & Validation targets \\
\midrule
Q1 (lowest) & 0--1,126 & 1,333,362 \\
Q2 & 1,127--6,491 & 1,343,753 \\
Q3 & 6,492--70,434 & 1,366,266 \\
Q4 (highest) & $\geq$70,435 & 1,325,257 \\
\bottomrule
\end{tabular}
\end{table}

For FineWebEdu, the frequency groups are shared within each model comparison. The following table reports the archived absolute NLL values for complete paired frequency evaluations.

\begin{table}[!htbp]
\centering
\footnotesize
\renewcommand{\arraystretch}{1.15}
\setlength{\tabcolsep}{3pt}
\caption{FineWebEdu validation NLL by token-frequency group.}
\label{tab:appendix-c-frequency-nll}
\begin{tabular}{@{}llrrrr@{}}
\toprule
Configuration & Architecture & Q1 NLL & Q2 NLL & Q3 NLL & Q4 NLL \\
\midrule
Coarse 137M & SwiGLU & 4.592255 & 3.647148 & 2.677261 & 1.420630 \\
Coarse 137M & ParGLU & 4.570064 & 3.623929 & 2.660884 & 1.415621 \\
Coarse 250M & SwiGLU & 4.282883 & 3.418670 & 2.541919 & 1.370355 \\
Coarse 250M & ParGLU & 4.269311 & 3.404256 & 2.533346 & 1.369546 \\
Coarse 395M & SwiGLU & 4.077668 & 3.268453 & 2.456620 & 1.330142 \\
Coarse 395M & ParGLU & 4.066733 & 3.262364 & 2.453851 & 1.325647 \\
Coarse 649M & SwiGLU & 3.918510 & 3.149591 & 2.386754 & 1.297781 \\
Coarse 649M & ParGLU & 3.915536 & 3.146249 & 2.382594 & 1.296380 \\
Balanced 253M & SwiGLU & 4.343970 & 3.464902 & 2.573867 & 1.377528 \\
Balanced 253M & ParGLU & 4.321065 & 3.447264 & 2.563732 & 1.370881 \\
Balanced 643M & SwiGLU & 3.979229 & 3.197044 & 2.416084 & 1.304732 \\
Balanced 643M & ParGLU & 3.972237 & 3.188614 & 2.411426 & 1.307638 \\
Fine 634M & SwiGLU & 4.004631 & 3.222734 & 2.435172 & 1.315456 \\
Fine 634M & ParGLU & 3.997384 & 3.220872 & 2.431648 & 1.314237 \\
Fine 1.03B & SwiGLU & 3.826524 & 3.097572 & 2.359416 & 1.285992 \\
Fine 1.03B & ParGLU & 3.811423 & 3.086894 & 2.353571 & 1.282918 \\
\bottomrule
\end{tabular}
\end{table}

\subsection{Predictive entropy}

Predictive entropy is computed as $-\sum_v p(v\mid x)\log p(v\mid x)$. Validation targets are grouped into quartiles using the predictive entropy of the SwiGLU baseline. The table reports the mean NLL of each architecture within the same quartile.

\begin{table}[!htbp]
\centering
\footnotesize
\renewcommand{\arraystretch}{1.15}
\setlength{\tabcolsep}{3pt}
\caption{FineWebEdu validation NLL by predictive-entropy quartile.}
\label{tab:appendix-c-entropy-nll}
\begin{tabular}{@{}llrrrr@{}}
\toprule
Configuration & Architecture & Q1 NLL & Q2 NLL & Q3 NLL & Q4 NLL \\
\midrule
Coarse 250M & SwiGLU & 0.623106 & 2.288614 & 3.586022 & 5.092047 \\
Coarse 250M & ParGLU & 0.625302 & 2.284753 & 3.573645 & 5.068868 \\
Coarse 395M & SwiGLU & 0.551943 & 2.153418 & 3.448767 & 4.955633 \\
Coarse 395M & ParGLU & 0.555016 & 2.152085 & 3.439644 & 4.938706 \\
Fine 251M & SwiGLU & 0.692927 & 2.406914 & 3.698900 & 5.185765 \\
Fine 251M & ParGLU & 0.693975 & 2.398451 & 3.681975 & 5.158156 \\
Fine 399M & SwiGLU & 0.587512 & 2.225235 & 3.516226 & 5.014837 \\
Fine 399M & ParGLU & 0.591172 & 2.224722 & 3.512444 & 5.003135 \\
Fine 634M & SwiGLU & 0.532798 & 2.114391 & 3.401931 & 4.905989 \\
Fine 634M & ParGLU & 0.535738 & 2.113440 & 3.397179 & 4.894983 \\
\bottomrule
\end{tabular}
\end{table}

\subsection{Predictive-entropy quartiles}

Predictive-entropy quartiles are defined using the SwiGLU baseline. Bootstrap confidence intervals use 4,095-target validation blocks and 10,000 repetitions.

\begin{table}[!htbp]
\centering
\scriptsize
\renewcommand{\arraystretch}{1.0}
\setlength{\tabcolsep}{2pt}
\caption{Predictive-entropy quartile results.}
\label{tab:appendix-c-entropy-bins}
\begin{tabular*}{\linewidth}{@{\extracolsep{\fill}}lrc c c r r c@{}}
\toprule
Configuration & Quartile & Targets & \shortstack{SwiGLU\\NLL / PPL} & \shortstack{ParGLU\\NLL / PPL} & $\Delta$NLL & Token win rate & \shortstack{95\% CI\\lower / upper} \\
\midrule
Balanced 402M & Q1 & 1342137 & \shortstack{0.581421\\1.7886} & \shortstack{0.583642\\1.7926} & 0.002221 & 46.31\% & \shortstack{0.001414\\0.003025} \\
Balanced 402M & Q2 & 1342136 & \shortstack{2.203195\\9.0539} & \shortstack{2.200020\\9.0252} & -0.003175 & 50.90\% & \shortstack{-0.004427\\-0.001940} \\
Balanced 402M & Q3 & 1342136 & \shortstack{3.493318\\32.8949} & \shortstack{3.481981\\32.5241} & -0.011337 & 51.32\% & \shortstack{-0.012842\\-0.009813} \\
Balanced 402M & Q4 & 1342136 & \shortstack{4.993000\\147.3780} & \shortstack{4.974964\\144.7436} & -0.018037 & 51.24\% & \shortstack{-0.019767\\-0.016356} \\
Fine 1.03B & Q1 & 1342137 & \shortstack{0.473886\\1.6062} & \shortstack{0.474691\\1.6075} & 0.000805 & 47.74\% & \shortstack{0.000091\\0.001500} \\
Fine 1.03B & Q2 & 1342136 & \shortstack{1.998558\\7.3784} & \shortstack{1.995006\\7.3522} & -0.003552 & 51.14\% & \shortstack{-0.004674\\-0.002410} \\
Fine 1.03B & Q3 & 1342136 & \shortstack{3.279083\\26.5514} & \shortstack{3.267240\\26.2388} & -0.011843 & 51.57\% & \shortstack{-0.013211\\-0.010487} \\
Fine 1.03B & Q4 & 1342136 & \shortstack{4.796093\\121.0366} & \shortstack{4.775925\\118.6200} & -0.020169 & 51.47\% & \shortstack{-0.021853\\-0.018520} \\
\bottomrule
\end{tabular*}
\end{table}

\subsection{Token-frequency by predictive-entropy cross analysis}

\begin{table}[!htbp]
\centering
\scriptsize
\renewcommand{\arraystretch}{1.0}
\setlength{\tabcolsep}{3pt}
\caption{Joint token-frequency and predictive-entropy $\Delta$NLL matrices.}
\label{tab:joint-delta-matrices}
\begin{tabular*}{0.92\linewidth}{@{\extracolsep{\fill}}lrrrr@{}}
\toprule
Configuration / entropy & Frequency Q1 & Q2 & Q3 & Q4\\
\midrule
\multicolumn{5}{l}{\textit{Balanced 402M}}\\
Entropy Q1 & 0.003089 & 0.001409 & 0.000394 & 0.003633\\
Entropy Q2 & -0.015641 & -0.016432 & 0.001771 & 0.004285\\
Entropy Q3 & -0.024307 & -0.017717 & -0.002863 & -0.001841\\
Entropy Q4 & -0.016238 & -0.016867 & -0.019058 & -0.027512\\
\addlinespace[2pt]
\multicolumn{5}{l}{\textit{Fine 1.03B}}\\
Entropy Q1 & -0.001353 & 0.001856 & 0.000695 & 0.001814\\
Entropy Q2 & -0.015651 & -0.009365 & -0.002219 & 0.002970\\
Entropy Q3 & -0.023047 & -0.014570 & -0.005902 & -0.005607\\
Entropy Q4 & -0.019013 & -0.018211 & -0.017716 & -0.036175\\
\bottomrule
\end{tabular*}
\end{table}

\begingroup
\scriptsize
\renewcommand{\arraystretch}{1.0}
\setlength{\tabcolsep}{1.3pt}
\begin{longtable}{@{}>{\raggedright\arraybackslash}p{0.92in}rrrc c r r r@{}}
\caption{Joint token-frequency and predictive-entropy auxiliary statistics.}\label{tab:joint-analysis-auxiliary}\\
\toprule
Configuration & Entropy & Frequency & Targets & \shortstack{SwiGLU\\NLL / PPL} & \shortstack{ParGLU\\NLL / PPL} & Token win rate & CI lower & CI upper\\
\midrule
\endfirsthead
\toprule
Configuration & Entropy & Frequency & Targets & \shortstack{SwiGLU\\NLL / PPL} & \shortstack{ParGLU\\NLL / PPL} & Token win rate & CI lower & CI upper\\
\midrule
\endhead
\midrule
\multicolumn{9}{r}{Continued on next page}\\
\endfoot
\bottomrule
\endlastfoot
Balanced 402M & Q1 & Q1 & 308480 & \shortstack{0.687767\\1.9893} & \shortstack{0.690856\\1.9954} & 45.75\% & 0.001086 & 0.005058\\
Balanced 402M & Q1 & Q2 & 324301 & \shortstack{0.589293\\1.8027} & \shortstack{0.590703\\1.8053} & 46.79\% & -0.000337 & 0.003182\\
Balanced 402M & Q1 & Q3 & 310484 & \shortstack{0.626087\\1.8703} & \shortstack{0.626481\\1.8710} & 46.18\% & -0.001186 & 0.002000\\
Balanced 402M & Q1 & Q4 & 398872 & \shortstack{0.458006\\1.5809} & \shortstack{0.461639\\1.5867} & 46.44\% & 0.002514 & 0.004816\\
Balanced 402M & Q2 & Q1 & 202631 & \shortstack{3.394260\\29.7926} & \shortstack{3.378619\\29.3302} & 51.80\% & -0.020097 & -0.011175\\
Balanced 402M & Q2 & Q2 & 244496 & \shortstack{2.882040\\17.8506} & \shortstack{2.865608\\17.5597} & 51.66\% & -0.019809 & -0.013012\\
Balanced 402M & Q2 & Q3 & 361863 & \shortstack{2.338353\\10.3642} & \shortstack{2.340124\\10.3825} & 50.34\% & -0.000580 & 0.004132\\
Balanced 402M & Q2 & Q4 & 533146 & \shortstack{1.347463\\3.8477} & \shortstack{1.351748\\3.8642} & 50.60\% & 0.002987 & 0.005583\\
Balanced 402M & Q3 & Q1 & 308948 & \shortstack{4.922255\\137.3119} & \shortstack{4.897949\\134.0146} & 51.84\% & -0.028577 & -0.020185\\
Balanced 402M & Q3 & Q2 & 341264 & \shortstack{4.073256\\58.7480} & \shortstack{4.055540\\57.7163} & 51.61\% & -0.020865 & -0.014505\\
Balanced 402M & Q3 & Q3 & 378597 & \shortstack{3.101933\\22.2409} & \shortstack{3.099070\\22.1773} & 50.65\% & -0.005116 & -0.000640\\
Balanced 402M & Q3 & Q4 & 313327 & \shortstack{1.925620\\6.8594} & \shortstack{1.923779\\6.8468} & 51.29\% & -0.003751 & 0.000150\\
Balanced 402M & Q4 & Q1 & 521947 & \shortstack{6.033467\\417.1589} & \shortstack{6.017229\\410.4397} & 51.03\% & -0.019442 & -0.013109\\
Balanced 402M & Q4 & Q2 & 431771 & \shortstack{5.024589\\152.1078} & \shortstack{5.007723\\149.5637} & 51.00\% & -0.019852 & -0.013968\\
Balanced 402M & Q4 & Q3 & 264517 & \shortstack{3.980022\\53.5182} & \shortstack{3.960964\\52.5079} & 51.51\% & -0.022058 & -0.016103\\
Balanced 402M & Q4 & Q4 & 123901 & \shortstack{2.662449\\14.3313} & \shortstack{2.634937\\13.9424} & 52.37\% & -0.031230 & -0.023755\\
Fine 1.03B & Q1 & Q1 & 323663 & \shortstack{0.549829\\1.7330} & \shortstack{0.548476\\1.7306} & 47.77\% & -0.003010 & 0.000322\\
Fine 1.03B & Q1 & Q2 & 332140 & \shortstack{0.473173\\1.6051} & \shortstack{0.475029\\1.6081} & 48.09\% & 0.000399 & 0.003339\\
Fine 1.03B & Q1 & Q3 & 306782 & \shortstack{0.508098\\1.6621} & \shortstack{0.508793\\1.6633} & 47.78\% & -0.000735 & 0.002069\\
Fine 1.03B & Q1 & Q4 & 379552 & \shortstack{0.382098\\1.4654} & \shortstack{0.383912\\1.4680} & 47.39\% & 0.000866 & 0.002753\\
Fine 1.03B & Q2 & Q1 & 208421 & \shortstack{2.964855\\19.3919} & \shortstack{2.949205\\19.0908} & 52.04\% & -0.019928 & -0.011402\\
Fine 1.03B & Q2 & Q2 & 245962 & \shortstack{2.561767\\12.9587} & \shortstack{2.552402\\12.8379} & 51.45\% & -0.012725 & -0.006012\\
Fine 1.03B & Q2 & Q3 & 354277 & \shortstack{2.154555\\8.6240} & \shortstack{2.152336\\8.6049} & 50.93\% & -0.004321 & -0.000130\\
Fine 1.03B & Q2 & Q4 & 533476 & \shortstack{1.257772\\3.5176} & \shortstack{1.260743\\3.5280} & 50.79\% & 0.001856 & 0.004074\\
Fine 1.03B & Q3 & Q1 & 300685 & \shortstack{4.577365\\97.2578} & \shortstack{4.554318\\95.0419} & 51.96\% & -0.026806 & -0.019237\\
Fine 1.03B & Q3 & Q2 & 336337 & \shortstack{3.837009\\46.3865} & \shortstack{3.822440\\45.7156} & 51.47\% & -0.017470 & -0.011682\\
Fine 1.03B & Q3 & Q3 & 379987 & \shortstack{2.968256\\19.4580} & \shortstack{2.962354\\19.3435} & 50.96\% & -0.008097 & -0.003700\\
Fine 1.03B & Q3 & Q4 & 325127 & \shortstack{1.864513\\6.4528} & \shortstack{1.858907\\6.4167} & 52.00\% & -0.007320 & -0.003834\\
Fine 1.03B & Q4 & Q1 & 509237 & \shortstack{5.818530\\336.4770} & \shortstack{5.799517\\330.1400} & 51.15\% & -0.022031 & -0.016023\\
Fine 1.03B & Q4 & Q2 & 427393 & \shortstack{4.863663\\129.4977} & \shortstack{4.845453\\127.1608} & 51.16\% & -0.021070 & -0.015416\\
Fine 1.03B & Q4 & Q3 & 274415 & \shortstack{3.850542\\47.0186} & \shortstack{3.832826\\46.1929} & 51.41\% & -0.020544 & -0.015015\\
Fine 1.03B & Q4 & Q4 & 131091 & \shortstack{2.583371\\13.2417} & \shortstack{2.547196\\12.7712} & 53.84\% & -0.039542 & -0.032851\\
\end{longtable}
\endgroup

\setcounter{table}{0}
\renewcommand{\thetable}{\Alph{section}\arabic{table}}
\renewcommand{\theHtable}{\Alph{section}\arabic{table}}
\section{Additional Fine-Tuning Results}

\subsection{Fine-tuning protocol}

The 1.03B SwiGLU and ParGLU checkpoints are independently fully fine-tuned on BillSum and PubMed. For each domain, 1,000 examples from the original training split form the validation set. BillSum uses two epochs (2,708 steps); PubMed uses one epoch (11,320 steps). The context length is 4,096 and the global batch comprises eight packed blocks. Loss is computed only over summary and EOS tokens. Document boundaries are isolated in attention and position indices are reset. AdamW is reinitialized with betas (0.9, 0.95), epsilon 1e-8, weight decay 0.01, gradient clipping 1.0, peak learning rate 2e-5, 5\% warmup, and cosine decay to 2e-6. Training uses BF16 autocast with FP32 parameters and updates the inherited dynamic routing bias. Final checkpoints are evaluated on the complete test sets under teacher forcing. Generation uses a fixed seed-42 subset of 256 examples per test set, greedy decoding and at most 256 new tokens.

\begin{table}[!htbp]
\centering
\scriptsize
\renewcommand{\arraystretch}{1.0}
\setlength{\tabcolsep}{2.5pt}
\caption{Teacher-forced evaluation for BillSum and PubMed.}
\label{tab:appendix-d-teacher-forced}
\begin{tabular*}{\linewidth}{@{\extracolsep{\fill}}llrrrr@{}}
\toprule
Test set & Checkpoint & NLL & PPL & Examples & Targets \\
\midrule
BillSum US & SwiGLU Base & 2.063899415 & 7.876624 & 3269 & 760056 \\
BillSum US & ParGLU Base & 2.006086902 & 7.434170 & 3269 & 760056 \\
BillSum US & SwiGLU SFT & 1.038612295 & 2.825294 & 3269 & 760056 \\
BillSum US & ParGLU SFT & 1.033575960 & 2.811100 & 3269 & 760056 \\
BillSum CA & SwiGLU Base & 1.921498584 & 6.831188 & 1237 & 471795 \\
BillSum CA & ParGLU Base & 1.901071607 & 6.693063 & 1237 & 471795 \\
BillSum CA & SwiGLU SFT & 1.859266189 & 6.419025 & 1237 & 471795 \\
BillSum CA & ParGLU SFT & 1.838985168 & 6.290152 & 1237 & 471795 \\
PubMed & SwiGLU Base & 2.527814 & 12.5261 & --- & --- \\
PubMed & ParGLU Base & 2.553693 & 12.8545 & --- & --- \\
PubMed & SwiGLU SFT Final & 1.702637 & 5.4884 & --- & --- \\
PubMed & ParGLU SFT Final & 1.696556 & 5.4551 & --- & --- \\
\bottomrule
\end{tabular*}
\end{table}

\begin{table}[!htbp]
\centering
\scriptsize
\renewcommand{\arraystretch}{1.0}
\setlength{\tabcolsep}{1.7pt}
\caption{Generation results for BillSum and PubMed.}
\label{tab:appendix-d-generation}
\begin{tabular*}{\linewidth}{@{\extracolsep{\fill}}llrrrrrrrr@{}}
\toprule
Test set & Checkpoint & ROUGE-1 & ROUGE-2 & ROUGE-L & \shortstack{Mean\\length} & \shortstack{Median\\length} & \shortstack{Hit\\cap} & \shortstack{EOS\\rate} & Repetition \\
\midrule
BillSum US & SwiGLU Base & 0.214860 & 0.048913 & 0.165728 & 256.00 & 256.0 & 1.000 & 0.000 & 0.757994 \\
BillSum US & ParGLU Base & 0.219128 & 0.053072 & 0.167748 & 256.00 & 256.0 & 1.000 & 0.000 & 0.742633 \\
BillSum US & SwiGLU SFT & 0.489320 & 0.306396 & 0.374692 & 194.43 & 235.5 & 0.4766 & 0.5234 & 0.322240 \\
BillSum US & ParGLU SFT & 0.496080 & 0.306581 & 0.380677 & 183.83 & 221.0 & 0.41797 & 0.58594 & 0.313307 \\
BillSum CA & SwiGLU Base & 0.279854 & 0.076381 & 0.194475 & 256.00 & 256.0 & 1.000 & 0.000 & 0.732451 \\
BillSum CA & ParGLU Base & 0.273060 & 0.067952 & 0.188739 & 256.00 & 256.0 & 1.000 & 0.000 & 0.738577 \\
BillSum CA & SwiGLU SFT & 0.348890 & 0.151433 & 0.221486 & 179.42 & 204.5 & 0.3984 & 0.6016 & 0.349196 \\
BillSum CA & ParGLU SFT & 0.391438 & 0.167235 & 0.244034 & 215.12 & 256.0 & 0.61328 & 0.39063 & 0.410882 \\
PubMed & SwiGLU Base & 0.223859 & 0.049160 & 0.178021 & 256.00 & --- & 1.000 & --- & 0.836535 \\
PubMed & ParGLU Base & 0.237524 & 0.056034 & 0.180365 & 256.00 & --- & 1.000 & --- & 0.797531 \\
PubMed & SwiGLU SFT Final & 0.380726 & 0.145092 & 0.235759 & 242.98 & --- & 0.8125 & --- & 0.401895 \\
PubMed & ParGLU SFT Final & 0.377447 & 0.144244 & 0.238927 & 240.65 & --- & 0.8242 & --- & 0.424998 \\
\bottomrule
\end{tabular*}
\par\smallskip
\begin{minipage}{\linewidth}\footnotesize\raggedright

\end{minipage}
\end{table}

\begin{table}[!htbp]
\centering
\footnotesize
\ParGLUTableStyle
\caption{Final validation results during fine-tuning.}
\label{tab:appendix-d-final-validation}
\begin{tabular}{@{}lrrrrrr@{}}
\toprule
Domain & Step & SwiGLU NLL & SwiGLU PPL & ParGLU NLL & ParGLU PPL & $\Delta$NLL \\
\midrule
BillSum & 2708 & 1.0213075 & 2.776823 & 1.0167314 & 2.764145 & $-0.0045761$ \\
PubMed & 11320 & 1.7040287 & 5.496045 & 1.6966049 & 5.455394 & $-0.0074238$ \\
\bottomrule
\end{tabular}
\end{table}
\section{Additional Sparsity Results}\label{app:additional-sparsity}

\subsection{Evaluation protocol}

We evaluate the pretrained Fine 1.03B models on 5,368,545 validation targets from 1,311 blocks. The backbone width is $C=896$ and the SwiGLU expert width is $H=576$, corresponding to $S=288$ and $R=432$ in ParGLU. The nine MoE layers contain 193,314,816 expert invocations in total. Both shared and routed experts are included. The dense first layer is unchanged. For SwiGLU, magnitude pruning retains the top-$K$ intermediate activations per token and activated expert before the down-projection. For ParGLU, exact zeros arise natively without pruning or sparsity regularization.

\subsection{Theoretical projection MACs}

A dense SwiGLU expert requires $3CH$ principal projection MACs per token. With $K$ retained activations,
\[
M_{\mathrm{Swi}}=2CH+CK,\qquad \eta_{\mathrm{Swi}}=\frac{H-K}{3H}.
\]

For standard ParGLU with ReLU$^2$ zero fraction $\rho$, neglecting the negligible zero fraction of its SwiGLU branch,
\[
M_{\mathrm{Par}}=2CS+CR+CS+C(1-\rho)R=3CH-\frac{3}{4}\rho CH.
\]

Consequently,
\[
\eta_{\mathrm{Par}}=\frac{\rho}{4}.
\]

All up-projections are computed in full. The savings correspond to skipped contributions in the down-projections. These expressions count principal projection MACs, not measured runtime or the overhead of sparse execution.

\textbf{Native activation sparsity.}

\begin{table}[!htbp]
\centering
\ParGLUTableStyle
\caption{Native activation sparsity and projection computation metrics.}
\label{tab:appendix-e-native-metrics}
\begin{tabular}{@{}lr@{}}
\toprule
Metric & Value \\
\midrule
Native NLL & 2.62819441 \\
Native PPL & 13.84874222 \\
ReLU$^2$ zero fraction & 85.7466\% \\
Shared ReLU$^2$ zero fraction & 83.8728\% \\
Routed ReLU$^2$ zero fraction & 86.3712\% \\
SwiGLU branch zero fraction & $\approx 0.00000118\%$ \\
Expert projection MAC saving & 21.4366\% \\
Full-model linear projection MAC saving & 9.6731\% \\
CPU evaluation time (s) & 4073.31 s \\
\bottomrule
\end{tabular}
\end{table}

\subsection{Magnitude pruning and effective projection widths}

\begin{table}[!htbp]
\centering
\scriptsize
\renewcommand{\arraystretch}{1.0}
\setlength{\tabcolsep}{0.5pt}
\caption{SwiGLU magnitude-pruning sweep and native ParGLU comparison.}
\label{tab:appendix-e-pruning}
\begin{tabular*}{\linewidth}{@{\extracolsep{\fill}}p{1.0in}rrrrrrrp{0.95in}@{}}
\toprule
Configuration & $K$ & NLL & PPL & $\Delta$NLL & $\Delta$PPL & \shortstack{Expert MAC\\saving} & \shortstack{Full-model MAC\\saving} & Device \\
\midrule
SwiGLU 0\% & 576 & 2.63688488 & 13.969619 & 0 & 0 & 0.0000\% & 0\% & CPU \\
SwiGLU 20\% & 461 & 2.63697676 & 13.970902 & +0.00009188 & +0.001284 & 6.6551\% & --- & CPU \\
SwiGLU 40\% & 346 & 2.63819226 & 13.987894 & +0.00130738 & +0.018276 & 13.3102\% & --- & CPU \\
SwiGLU 50\% & 288 & 2.64045676 & 14.019606 & +0.00357189 & +0.049987 & 16.6667\% & --- & CPU \\
SwiGLU 64.5\% & 204 & 2.65147582 & 14.174943 & +0.01459094 & +0.200532 & 21.5278\% & $\approx$9.7142\% & CPU \\
SwiGLU 70\% & 173 & 2.66155952 & 14.318602 & +0.02467464 & +0.348983 & 23.3218\% & --- & CPU 594 + GPU 717 blocks \\
SwiGLU 75\% & 144 & 2.67787416 & 14.554121 & +0.04098928 & +0.584502 & 25.0000\% & --- & GPU \\
SwiGLU 80\% & 115 & 2.70560027 & 14.963296 & +0.06871539 & +0.993677 & 26.6782\% & --- & GPU \\
Native ParGLU & --- & 2.62819441 & 13.84874222 & --- & --- & 21.4366\% & 9.6731\% & CPU \\
\bottomrule
\end{tabular*}
\end{table}
\textbf{Effective projection widths.}

\begin{table}[!htbp]
\centering
\ParGLUTableStyle
\caption{Effective projection widths under dense and sparse computation.}
\label{tab:appendix-e-effective-widths}
\begin{tabular}{@{}lrrrr@{}}
\toprule
Model & Up/gate width & Effective down width & Total effective width & PPL \\
\midrule
SwiGLU K576 & 1152 & 576 & 1728 & 13.969619 \\
Dense ParGLU & 1008 & 720 & 1728 & 13.84874222 \\
SwiGLU K346 & 1152 & 346 & 1498 & 13.987894 \\
SwiGLU K204 & 1152 & 204 & 1356 & 14.174943 \\
Native ParGLU & 1008 & 349.57 & 1357.57 & 13.84874222 \\
\bottomrule
\end{tabular}
\end{table}

The unpruned baseline, the near-equal-cost $K=204$ baseline, and native ParGLU are evaluated on CPU. The 70\% pruning result combines 594 CPU blocks and 717 GPU blocks; the 75\% and 80\% results use GPU. All rows cover the same number of validation targets. The dense and native ParGLU rows refer to the same model.

\setcounter{table}{0}
\renewcommand{\thetable}{\Alph{section}\arabic{table}}
\renewcommand{\theHtable}{\Alph{section}\arabic{table}}
\section{Research Origin}

ParGLU did not begin as a carefully planned research project. I was running other experiments when my GPU happened to be idle. With some spare computing time, I had a random thought: what would happen if I added a parallel C-to-C FFN branch to SwiGLU? It just seemed interesting, so I decided to try it.

My first attempt combined SwiGLU with GELU in dense language models. The early results looked promising, but they were not consistent. During these experiments, I began thinking about the nonlinear structures themselves. SwiGLU has a leading second-order interaction, while GELU contains a first-order term. ReLU$^2$ seemed like a more natural partner. So I replaced GELU with ReLU$^2$.

The dense experiments still did not give me a clear answer. Some configurations improved, while others did not. I then decided to try the same idea in Mixture-of-Experts models. To my surprise, the results became much more consistent. What began as a small experiment gradually turned into the work presented in this paper.

Even now, I still don’t quite understand why something so simple actually works.

Sometimes, that is enough to begin.

\end{document}